%% file: neurips_2026.tex
\documentclass{article}

\usepackage[main, final]{neurips_2026}
\usepackage[utf8]{inputenc} % allow utf-8 input
\usepackage[T1]{fontenc}    % use 8-bit T1 fonts
\usepackage{hyperref}       % hyperlinks
\usepackage{url}            % simple URL typesetting
\usepackage{booktabs}       % professional-quality tables
\usepackage{amsfonts}       % blackboard math symbols
\usepackage{nicefrac}       % compact symbols for 1/2, etc.
\usepackage{microtype}      % microtypography
\usepackage{xcolor}         % colors
\usepackage{enumitem}
\usepackage{graphicx}
\usepackage{multirow}
\usepackage{algorithm}
\usepackage{algorithmic}
\usepackage{amsmath}
\usepackage{cleveref}
\usepackage{wrapfig}
\usepackage[normalem]{ulem}
\usepackage{etoolbox}
\usepackage{xparse}

\newif\ifshowrev
\showrevtrue   % <-- change to \showrevfalse for the final upload
\ifdefined\finalclean\showrevfalse\fi
\ifshowrev
  
  \newcommand{\del}[1]{{\color{red}\sout{#1}}}
  
\else
  
  \newcommand{\del}[1]{\unskip}
  
\fi
\AtBeginDocument{%
  \let\saftorigbibitem\bibitem
  \RenewDocumentCommand\bibitem{om}{%
    \IfValueTF{#1}{\saftorigbibitem[#1]{#2}}{\saftorigbibitem{#2}}%
    \normalcolor\ifshowrev\ifcsdef{saftrevbib@#2}{\color{red}}{}\fi}}
\title{Sparse Fine-Tuning for Parameter-Efficient Adversarial Training}

\author{%
  Zhaoxin Wang \\
  School of Artificial Intelligence \\
  Xidian University \\
  Xi'an, China
  \And
  Zihang Ding \\
  School of Artificial Intelligence \\
  Xidian University \\
  Xi'an, China
  \And
  Handing Wang\thanks{%
    \begin{tabular}[t]{@{}l@{}}
      Corresponding author. 
      Email: \texttt{wangzhaoxin@stu.xidian.edu.cn}, \texttt{hdwang@xidian.edu.cn}.
    \end{tabular}} \\
  School of Artificial Intelligence \\
  Xidian University \\
  Xi'an, China
}

\begin{document}

\maketitle

\vspace{-10pt}
\begin{abstract}
Deep neural networks achieve strong performance across many tasks but remain vulnerable to adversarial perturbations. Adversarial training (AT) is one of the most effective defenses, yet it suffers from high memory cost on large models and the persistent clean-robust accuracy trade-off. Recent work introduces parameter-efficient fine-tuning, such as LoRA, into AT, but this constraint induces a notable robustness gap relative to full-parameter AT. In this work, we revisit parameter-efficient adversarial fine-tuning from a parameter space perspective. Through gradient analysis, we find that adversarial optimization is highly concentrated on a small subset of parameters, which we call Robustness-Critical Parameters (RCPs), suggesting that robustness is encoded in a sparse subspace. Building on this observation, we propose Sparse Adversarial Fine-Tuning (SAFT), which identifies RCPs via adversarial gradient saliency and performs AT by updating only these parameters while freezing the rest. Across architectures and datasets, SAFT outperforms LoRA-based methods in both clean and robust accuracy with fewer trainable parameters, and approaches or surpasses full parameter AT in robustness while updating only about 5\% of parameters.
\end{abstract}

\section{Introduction}
Deep learning has achieved excellent performance across a wide range of tasks \citep{lecun2015deep}, such as image classification 
\cite{he2020deep}, semantic segmentation \cite{hao2020brief}, and object detection \cite{jiao2019survey}. However, deep learning models are vulnerable to adversarial attacks, in which imperceptible perturbations added to the input can lead to wrong predictions \cite{szegedy2013intriguing}. To improve the model's robustness against these attacks, one of the most effective defenses is adversarial training (AT) \cite{madry2017towards}. AT can be formulated as a bi-level optimization problem, where the inner maximization generates adversarial perturbations, typically using a multi-step procedure called projected gradient descent (PGD)\cite{madry2017towards}, and the outer minimization optimizes the model parameters on the perturbed data.

Although AT significantly improves the robustness of the deep learning models, AT suffers from two longstanding issues. First, the bi-level optimization in AT requires multiple forward and backward passes per training step to generate adversarial perturbations, leading to substantially higher memory and time costs than standard training. Second, as adversarial examples and clean examples can be viewed as samples from two different distributions, AT induces a clean-robust accuracy trade-off that models trained to be robust often suffer a notable accuracy degradation on benign samples.

To reduce the computational overhead of AT, one line accelerates the 
inner perturbation generation process \cite{zhang2019you, 
shafahi2019adversarial, wang2024efficient, zheng2020efficient}. FastAT 
\cite{wong2020fast} replaces multi-step PGD with single-step FGSM, but 
it suffers from catastrophic overfitting, which subsequent studies seek 
to mitigate \cite{andriushchenko2020understanding, zhang2022revisiting, 
jia2024improving, wang2024preventing}. While these methods reduce 
training time, they update all model parameters and leave the memory 
bottleneck unaddressed. To reduce memory cost on large transformer-based 
models, recent work introduces parameter-efficient fine-tuning (PEFT) 
into AT. LoFT \cite{LoFT}, LNLoRA \cite{yuan2025fulllora}, and AdvLoRA 
\cite{ji2025enhancing} all build on LoRA \cite{hu2022lora} to limit 
trainable parameters to a low-rank subspace. However, LoRA-based PEFT-AT 
exhibits a persistent robustness gap relative to full-parameter AT 
\cite{huang2024allora, chang2025lora}, suggesting that the low-rank 
constraint may be misaligned with the optimization 
directions needed for robustness.

Unlike these efficiency methods, another line of work targets the trade-off problems~\cite{zhang2019theoretically, wang2023adversarial, wasedarethinking}. TRADES \cite{zhang2019theoretically} reformulates the AT objective to balance clean and robust losses through a tunable regularizer. MART \cite{wang2019improving} reweights misclassified examples to better handle hard cases. Carmon et al. \cite{carmon2019unlabeled} leverage additional unlabeled data to push the trade-off frontier outward. While these methods improve the trade-off frontier, they all operate at the loss or data level and implicitly treat the trade-off as a property shared uniformly across all parameters. Whether the trade-off has structure in parameter space, and whether it can be mitigated by acting on that structure, remains largely unexplored.

In this work, we revisit AT from a parameter-space perspective and ask whether the two above issues share a common origin. Through gradient analysis, we observe that adversarial optimization is concentrated on a small subset of parameters. A small fraction of parameters, about 5\%, carry the majority of the cumulative gradient mass, while over 90\% remain near zero throughout training. These observations suggest that robustness is not uniformly distributed across the network but is encoded in a sparse subset of parameters, which we refer to as Robustness Critical Parameters (RCPs).

This sparsity has two important implications. From an efficiency perspective, full-parameter AT is over-parameterized for robustness, and updating only the RCPs should suffice to recover most of the robustness gain. From a trade-off perspective, if the adversarial signal lives in a sparse subspace of parameters, restricting updates to that subspace leaves all remaining pretrained weights unchanged. The two longstanding issues of AT may therefore be approached jointly, not by modifying the loss or expanding the data, but by acting on the parameter-space structure of robustness itself.

Building on this, we propose Sparse Adversarial Fine-Tuning (SAFT), a parameter-efficient adversarial fine-tuning framework that identifies 
RCPs via adversarial gradient saliency and performs adversarial fine-tuning by updating only these parameters while freezing the rest. The identification is performed once before fine-tuning and requires no additional learnable importance scores. SAFT is orthogonal to existing AT methods and can be easily integrated with different AT objectives. Our contribution can be summarised as:

\begin{itemize}[leftmargin=1.5em]
    \item We identify a sparse parameter-space structure underlying adversarial robustness, where adversarial optimization is concentrated on a small subset of parameters, which we call RCPs. This concentration holds across datasets and on a convolutional backbone. The RCPs are highly consistent across different attack types, indicating that they reflect an intrinsic property of the model rather than an artifact of any specific attack.
    \item We propose SAFT, a sparse parameter-efficient adversarial fine-tuning framework that updates only the RCPs while freezing the rest of the model. By confining adversarial updates to the RCPs, SAFT leaves the remaining pretrained weights unchanged and directs robustness gains only to where they are needed. It reduces optimizer-state, persistent-gradient and peak training memory relative to full-parameter AT.
    \item Across architectures and datasets, SAFT outperforms LoRA-based adversarial fine-tuning and approaches or surpasses full-parameter adversarial fine-tuning in robust accuracy while updating only about 5\% of parameters. The localization of robustness in the RCPs explains why SAFT improves the clean-robust trade-off frontier, offering a new parameter-space perspective. Control experiments show that RCPs are robustness-specific rather than generally important parameters.
\end{itemize}

\section{Related Work}
\vspace{-5pt}
\subsection{Efficient Adversarial Training}
\vspace{-5pt}
A major limitation of AT is the high computational cost incurred by multi-step adversarial example generation. Existing efficient AT methods mainly reduce the cost of the inner maximization. FastAT~\cite{wong2020fast} replaces multi-step PGD with single-step FGSM~\cite{goodfellow2014explaining}, but suffers from catastrophic overfitting. Subsequent methods improve the stability of single-step AT through gradient regularization or modified perturbation updates, such as GradAlign~\cite{andriushchenko2020understanding} and FGSM-PCO~\cite{wang2024preventing}. Other methods reduce computation by reusing gradients or perturbations. FreeAT~\cite{shafahi2019adversarial} combines adversarial example generation and model update within the same backward pass. YOPO~\cite{zhang2019you} avoids full network backpropagation during inner maximization by using an approximation based on the first layer.

Recently, PEFT has been introduced to reduce the memory cost of adversarial adaptation of transformer-based large models. LoRA-based methods restrict trainable parameters to low-rank adaptation modules. Yuan et al.~\cite{yuan2025fulllora} apply LoRA to most modules and introduce LayerNorm before LoRA modules to improve robustness, while AdvLoRA~\cite{ji2025enhancing} improves LoRA-based adversarial adaptation through reparameterization. DoRA~\cite{liu2024dora} decomposes each pretrained weight into a magnitude and a direction and applies the low-rank update to the direction.

\subsection{Accuracy-Robustness Trade-off}
The trade-off between clean accuracy and adversarial robustness is hard to handle in AT. Improving robustness against adversarial perturbations often comes at the cost of lower accuracy on benign examples. Prior studies have provided several explanations for this phenomenon. Tsipras et al.~\cite{tsiprasrobustness} show that robustness and accuracy can be inherently conflicting under certain data distributions. Ilyas et al.~\cite{ilyas2019adversarial} attribute this conflict to non-robust but predictive features, which are useful for clean classification but sensitive to small perturbations. These studies suggest that the trade-off is closely related to the features learned by the model. To mitigate this issue, TRADES~\cite{zhang2019theoretically} introduces a regularized objective that balances natural accuracy and boundary robustness. MART~\cite{wang2019improving} improves robust generalization by reweighting misclassified examples. Semi-supervised AT and robust self-training leverage additional unlabeled data to push the trade-off frontier outward~\cite{carmon2019unlabeled}. Recent methods further explore invariance regularization, input-level reweighting, attention regularization, and label-space reformulation to reduce the conflict between clean and adversarial learning~\cite{wasedarethinking, zhangimproving}.

\subsection{Parameter-Space Sparsity in Robustness}
Parameter selection has been widely studied as a way to improve training efficiency while preserving model performance. Selective PEFT methods update only a subset of pretrained parameters instead of introducing additional adapter modules. Diff pruning~\cite{guo2021parameter} learns a sparse mask over parameter differences to identify task-relevant updates. PaFi~\cite{liao2023parameter} fine-tunes parameters with small absolute magnitudes in pretrained models. LT-SFT~\cite{ansell2022composable} selects parameters that change the most during the first epoch of fine-tuning. Parameter sparsity is also closely related to pruning and subnetwork discovery. The lottery ticket hypothesis~\cite{frankle2018the} shows that dense neural networks contain sparse subnetworks that can be trained in isolation to match the performance of the original network. A large body of work on structured pruning further shows that many parameters or components can be removed while retaining accuracy~\cite{cheng2024survey, zhu2025comprehensive, vadera2022methods}. These studies suggest that model capabilities are not uniformly distributed across all parameters, but can be concentrated in a subset of critical components.

\section{Empirical Findings}
\label{sec:empirical_findings}
Most existing AT methods update all model parameters during optimization. This practice assumes that adversarial robustness is distributed across the parameter space. However, our empirical analysis suggests a different picture. Robust adaptation is highly sparse, with adversarial update signals concentrated in a small subset of parameters. In this section, we present motivating evidence based on AT of a pretrained ViT-B model~\cite{dosovitskiy2020image} on CIFAR-10~\cite{krizhevsky2009learning}.

\subsection{Adversarial Gradients Concentrate on Few Parameters}
We analyze how adversarial optimization is distributed across 
model parameters. During AT, we accumulate the gradient saliency of each parameter under the adversarial loss and normalize the values for comparison. This measures how strongly each parameter participates in adversarial optimization throughout training.

\begin{wrapfigure}{l}{0.4\textwidth}
\centering
\setlength{\abovecaptionskip}{2pt}
\setlength{\belowcaptionskip}{0pt}
\includegraphics[width=0.39\textwidth]{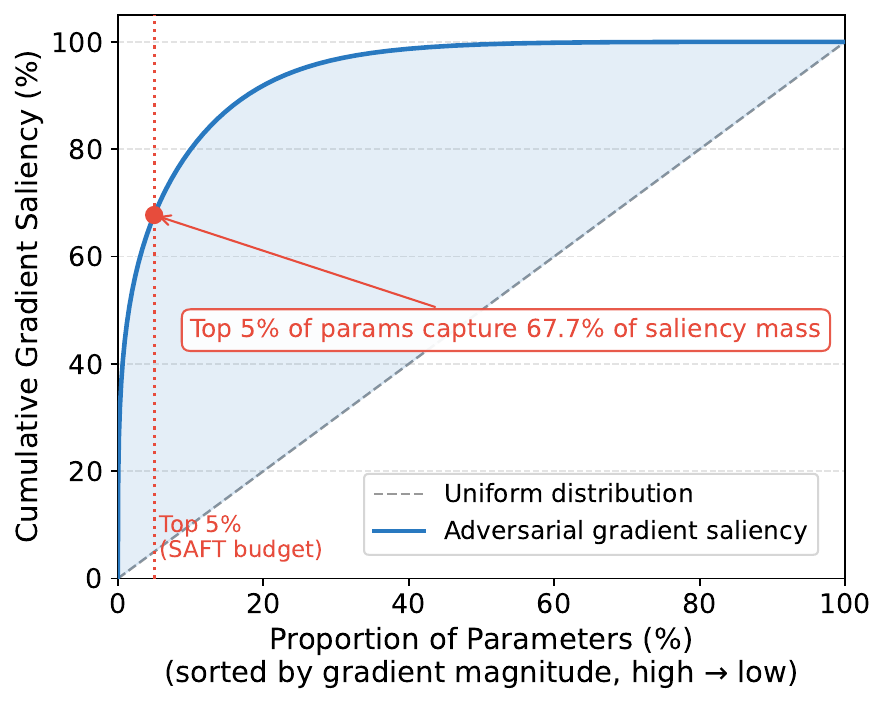}
\caption{Top 5\% of ViT-B parameters capture 67.7\% of adversarial gradient.}
\label{fig:abs_gradient}
\vspace{-28pt}
\end{wrapfigure}

As shown in Fig.~\ref{fig:abs_gradient}, the cumulative distribution is highly skewed. The top 5\% of parameters capture 67.7\% of the total saliency mass, far above the 5\% expected under a uniform distribution. The remaining 95\% of parameters together account for only about a third of the cumulative saliency, with the bottom half contributing almost nothing. This finding suggests that adversarial fine-tuning does not update all parameters uniformly. Instead, the adversarial optimization signal is concentrated in a sparse subset of parameters, while the vast majority receive negligible adversarial gradients. From this perspective, full-parameter adversarial fine-tuning is over-parameterized for the purpose of robust adaptation.

\subsection{High-Gradient Parameters Are Critical for the Robust Model}
To examine whether high-gradient parameters are functionally important for the trained model, we perform a parameter intervention experiment. Starting from an adversarially fine-tuned ViT-B model on CIFAR-10, we identify the parameters with the largest accumulated adversarial gradients and mask them by setting their values to zero. We compare this intervention to masking a random subset, and re-evaluate both clean and robust accuracy without further training.

\begin{wraptable}{r}{0.4\textwidth}
\centering
\vspace{-12pt}
\caption{Masking RCPs is more destructive than masking random parameters.}
\label{tab:mask}
\footnotesize
\setlength{\tabcolsep}{6pt}
\renewcommand{\arraystretch}{1.1}
\begin{tabular}{ccc}
\toprule
Setting & Clean & PGD-20 \\
\midrule
Baseline & 85.04 & 43.60 \\
\midrule
5\% Random& 81.83 & 39.48 \\
\textbf{RCPs 5\%} & \textbf{29.82} & \textbf{3.70} \\
\midrule
Random 10\% & 66.46 & 32.18 \\
\textbf{RCPs 10\%} & \textbf{24.35} & \textbf{4.75} \\
\bottomrule
\end{tabular}
\end{wraptable}

The results are shown in Table~\ref{tab:mask}. Masking the top 5\% high-gradient parameters causes robust accuracy to collapse from 43.60\% to 3.70\%, a relative loss of more than 90\%, while masking a random 5\% subset reduces it only to 39.48\%. The destructive effect depends on which parameters are removed, not how many. Clean accuracy is also affected by masking high-gradient parameters, indicating that these parameters are critical to the overall function of the network rather than to robustness alone. However, the relative loss in robust accuracy is substantially larger than that in clean accuracy, showing that the robust behavior of the trained model relies on this small subset.

Together with the gradient distribution in Fig.~\ref{fig:abs_gradient}, this intervention indicates that high-gradient parameters are both statistically salient and causally critical for the robust behavior of the adversarially trained model. We refer to these parameters as RCPs. The above findings motivate a sparse view of AT. If only a small subset of parameters receives strong adversarial update signals and is critical for robustness, updating the entire model may be unnecessary. Based on these observations, we propose SAFT, which first identifies robustness-critical parameters using adversarial gradient saliency and then updates only these parameters during adversarial fine-tuning. By concentrating adversarial optimization on RCPs while freezing the remaining parameters, SAFT targets the sparse subspace where adversarial optimization lives, and leaves the remaining pretrained weights unchanged.

\section{Method}
\label{sec:method}
Motivated by the empirical findings in Section~\ref{sec:empirical_findings}, 
we propose SAFT. Instead of updating all parameters during adversarial fine-tuning, SAFT first identifies RCPs using adversarial gradient saliency, and then updates only these parameters while freezing the rest of the model. This design preserves the original model architecture, introduces no adapter modules, and directly confines adversarial adaptation to the parameter subset most relevant to robustness.

\subsection{Formulation}
\label{sec:method_overview}

Let $f_\theta$ denote a pretrained model with parameters $\theta$, 
and let $\mathcal{D}$ denote the downstream training distribution. 
Standard AT optimizes
\begin{equation}
   \min_{\theta}\;
\mathbb{E}_{(x,y)\sim\mathcal{D}}
\left[\,
\max_{\|\delta\|_\infty \leq \epsilon}\;
\ell\bigl(f_\theta(x+\delta),\,y\bigr)
\right], 
\end{equation}

where $\delta$ is the adversarial perturbation and $\epsilon$ is the 
perturbation budget. Existing AT methods update the full parameter 
vector $\theta$ during the outer minimization. SAFT instead introduces 
a binary update mask $m\in\{0,1\}^{|\theta|}$, where $m_i=1$ indicates 
that $\theta_i$ is trainable and $m_i=0$ indicates that it is frozen. 
Given an AT loss $\mathcal{L}_{adv}$, the masked 
update is
\begin{equation}
\theta^{t+1}
=
\theta^{t}
-
\eta\,\bigl(m\odot\nabla_\theta\mathcal{L}_{adv}(\theta^{t})\bigr),
\label{eq:masked_update}
\end{equation}
where $\eta$ is the learning rate and $\odot$ denotes element-wise 
multiplication. The central design question is how to construct $m$ 
so that the selected parameters are those most responsible for robust 
adaptation. SAFT constructs this mask using adversarial gradient 
saliency.

\subsection{Robustness-Critical Parameter Identification}
\label{sec:rcp_identification}

We identify RCPs through a short calibration phase before sparse 
fine-tuning. Starting from the pretrained model $f_{\theta}$, we run 
a small number of AT steps on a calibration set $\mathcal{D}_{cal}$ 
and accumulate the adversarial gradient magnitude of each parameter 
throughout this phase. For parameter $\theta_i$, the accumulated 
saliency score is
\begin{equation}
s_i
=
\mathbb{E}_{(x,y)\sim\mathcal{D}_{cal}}
\left[\,
\bigl|
\nabla_{\theta_i}\,
\ell\bigl(f_{\theta}(x+\delta), y\bigr)
\bigr|
\,\right],
\end{equation}
where $\delta$ is the adversarial perturbation generated against the 
current model by the same attack used for AT, 
\begin{equation}
\delta
=
\arg\max_{\|\delta'\|_\infty\leq\epsilon}\;
\ell\bigl(f_{\theta}(x+\delta'), y\bigr).
\end{equation}

The expectation is taken over the calibration data. The saliency score $s_i$ measures how strongly parameter $\theta_i$ participates in adversarial optimization across the calibration phase. Parameters with larger scores receive stronger adversarial update signals and are therefore more likely to be critical for robust adaptation. We select the top $\rho$ fraction of parameters according to $s_i$ as RCPs:
\begin{equation}
\mathcal{R}
=
\operatorname{Top}_{\rho}\bigl(\{s_i\}_{i=1}^{|\theta|}\bigr),
\quad
m_i =
\begin{cases}
1, & \theta_i\in\mathcal{R},\\
0, & \theta_i\notin\mathcal{R}.
\end{cases}
\end{equation}

The mask is computed once during the calibration phase and fixed throughout AT. In all experiments, the calibration phase is a single epoch over the training set. When applying SAFT to transformer-based models, we restrict the search space to the main transformation modules, following common practice in parameter-efficient fine-tuning. Specifically, we search for RCPs in the projection matrices of self-attention modules ($W_Q$, $W_K$, $W_V$, $W_O$) and in the two linear layers of the feed-forward networks ($W_{up}$, $W_{down}$). This focuses the update budget on modules that dominate representation transformation.

\subsection{Sparse Adversarial Fine-Tuning}
\label{sec:saft_training}

After constructing the RCPs mask, SAFT performs AT with masked gradients according to the update rule introduced in Section~\ref{sec:method_overview}. Only RCPs are updated, while all other parameters remain at their pretrained values. SAFT modifies only the update space, not the adversarial objective itself, so it can be combined with different AT losses such as PGD-AT~\cite{madry2017towards}, TRADES~\cite{zhang2019theoretically}, and MART~\cite{wang2019improving}.

SAFT differs from both pruning and LoRA-based PEFT. Unlike pruning, 
SAFT does not remove parameters or alter the inference architecture. 
The full pretrained model is retained, and the mask only controls 
training-time updates. Unlike LoRA-based methods, SAFT does not 
introduce low-rank adapters or constrain updates to an external 
low-rank subspace. Instead, it directly updates a sparse subset of 
the original model parameters selected by adversarial gradient 
saliency, providing a sparse counterpart to LoRA-based PEFT AT.

This sparse update strategy has two benefits. First, it reduces training memory, since persistent gradients and optimizer states need to be maintained only for the selected $\rho$ fraction of parameters. The backward pass still computes the full gradient in Eq.~\eqref{eq:masked_update}, so SAFT does not reduce the FLOPs of AT but only the selected coordinates keep persistent gradients and optimizer states. Second, by confining adversarial adaptation to RCPs and keeping all other parameters at their pretrained values, SAFT minimizes the unnecessary perturbation that full-parameter AT introduces across the network. The remaining parameters of the model stay close to its pretrained state, which achieves a better trade-off between training cost and robust adaptation.

\section{Experimental Results}
\label{sec:experiments}
\subsection{Experimental Setup}

\paragraph{Datasets and Models.} 
We conduct experiments across datasets. CIFAR-10 and CIFAR-100~\cite{krizhevsky2009learning} each contain 60,000 images of size $32\times32$, with 50,000 for training and 10,000 for testing, distributed across 10 and 100 classes respectively. TinyImageNet~\cite{le2015tiny} contains 200 classes with 500 training images and 50 for testing per class. We evaluate SAFT on three pretrained backbones, including ViT-B, ViT-S~\cite{dosovitskiy2020image}, and Swin-B~\cite{liu2021swin}, and additionally ResNet-50~\cite{he2016resnet} in Appendix~\ref{app:resnet}. ViT-B and ViT-S both contain 12 Transformer blocks, with hidden widths of 768 and 384, while Swin-B adopts a hierarchical architecture with a shifted-window 
attention mechanism. All models are pretrained on ImageNet-1K~\cite{russakovsky2015imagenet}. We keep the default architectures and adjust the image and patch sizes to match the 
input resolution. For CIFAR-10 and CIFAR-100, the image size is set to 32 and the patch size to 4. For TinyImageNet, the image size is set to 64 and the patch size to 8.

\paragraph{Evaluation Metrics.} 
We report classification accuracy on clean inputs and adversarial examples. Four attacks, PGD-20~\cite{madry2017towards}, CW-20~\cite{7958570}, PGD-100~\cite{madry2017towards}, and AutoAttack~\cite{croce2020reliable} are used to generate perturbations, where the number following PGD or CW refers to the number of iterations. 

\paragraph{Comparison Methods.} 
We compare SAFT against full-parameter adversarial fine-tuning and two LoRA-based PEFT-AT methods, LoRA~\cite{hu2022lora} and LNLoRA~\cite{yuan2025fulllora}. To verify the effectiveness of adversarial gradient saliency for parameter selection, we compare against random parameter selection, weight-magnitude selection and clean-data gradient saliency under the same parameter budget.

\paragraph{Training Details.} 
We use SGD~\cite{robbins1951stochastic} with momentum 0.9 and weight decay $10^{-4}$. Models are trained for 40 epochs with an initial learning rate of 0.04, decayed by a factor of 10 at epochs 35 and 38. For adversarial perturbations, we use the $\ell_\infty$ norm with $\epsilon = 8/255$ and step size $\alpha = 2/255$. For LoRA, LNLoRA and DoRA, the rank is set to $r = 32$. For SAFT, we select the top $5\%$ of parameters as RCPs. All experiments are conducted on a single NVIDIA RTX 4090 GPU.

\subsection{Main Results across Architectures}
\label{sec:main_arch}

We compare SAFT against full-parameter adversarial fine-tuning and two LoRA-based PEFT-AT methods on three pretrained backbones. Results on CIFAR-10 
are reported in Table~\ref{tab:models}. SAFT consistently outperforms LoRA-based PEFT-AT across all architectures. On every backbone and under every attack, SAFT achieves higher clean and robust accuracy than both LoRA and LNLoRA, while using fewer trainable parameters. The improvement is substantial. On ViT-B, SAFT improves over LoRA by 3.04\% in clean accuracy and 2.83\% on PGD-20, while reducing trainable parameters from 4.7M to 3.1M. On ViT-S, the gain is even larger, with SAFT improving robust accuracy by 2.99\% on PGD-20 and 3.50\% on AutoAttack. On Swin-B, SAFT improves over LoRA by 3.67\% on PGD-20 and 3.35\% on AutoAttack, with comparable clean accuracy.

\begin{table}[h]
\centering
\caption{Comparison across architectures on CIFAR-10. Trainable parameters refer to updated parameters during AT. Below each backbone: its total parameter count, to which LoRA-based methods add adapters.}
\label{tab:models}
\setlength{\tabcolsep}{3.9pt}
\renewcommand{\arraystretch}{1.0}
\begin{tabular}{llcccccc}
\toprule
Model & Method & Trainable Params & Clean & PGD-20 & CW-20 & PGD-100 & AutoAttack \\
\midrule
\multirow{5}{*}{\shortstack[l]{ViT-B\\(85.2M)}}
 & Pretrained      & --     & 96.26 & 0.00  & 0.00  & 0.00  & 0.00  \\
 & Full Parameter  & 85.2M (100\%)  & 85.04 & 43.60 & 44.57 & 42.96 & 41.16 \\
 \cmidrule(l){2-8}
 & LoRA            & 4.7M (5.5\%)   & 79.51  & 41.58 & 41.27 & 41.31 & 38.63 \\
 & LNLoRA          & 4.8M (5.6\%)   & 79.59 & 41.79 & 41.50 & 41.40 & 39.07 \\
  
 & \textbf{SAFT}   & \textbf{3.1M} (3.6\%) & \textbf{82.55} & \textbf{44.41} & \textbf{44.87} & \textbf{44.04} & \textbf{41.61} \\
\midrule
\multirow{5}{*}{\shortstack[l]{ViT-S\\(21.4M)}}
 & Pretrained      & --     & 94.49 & 0.00  & 0.00  & 0.00  & 0.00  \\
 & Full Parameter  & 21.4M (100\%)  & 83.09 & 42.60 & 43.96 & 42.19 & 40.53 \\
 \cmidrule(l){2-8}
 & LoRA            & 1.1M (5.1\%)   & 75.57 & 38.50 & 37.73 & 38.10 & 34.97 \\
  & LNLoRA          & 1.1M (5.1\%)   & 75.09 & 38.47 & 37.62 & 38.43 & 35.36 \\
 & \textbf{SAFT}   & \textbf{0.8M} (3.7\%) & \textbf{78.13} & \textbf{41.49} & \textbf{41.43} & \textbf{41.11} & \textbf{38.47} \\
\midrule
\multirow{5}{*}{\shortstack[l]{Swin-B\\(85.9M)}}
 & Pretrained      & --    & 87.26 & 0.00  & 0.00  & 0.00  & 0.00  \\
 & Full Parameter  & 85.9M (100\%)  & 72.00 & 39.03 & 38.24 & 38.80 & 35.92 \\
 \cmidrule(l){2-8}
 & LoRA            & 5.9M (6.9\%)   & 67.29 & 33.38 & 32.07 & 33.15 & 29.85 \\
  & LNLoRA          & 5.9M (6.9\%)   & 67.30 & 33.42 & 32.09 & 33.16 & 29.84 \\
 & \textbf{SAFT}   & \textbf{4.1M} (4.8\%) & \textbf{67.37} & \textbf{37.05} & \textbf{35.75} & \textbf{36.53} & \textbf{33.20} \\
\bottomrule
\end{tabular}
\end{table}

SAFT approaches or surpasses full-parameter AT in robust accuracy while updating fewer than 5\% of parameters. On ViT-B, SAFT surpasses Full AT under all four attacks, despite training only 3.6\% of the parameters. On ViT-S and Swin-B, SAFT approaches Full AT in robustness, with a remaining gap of about 1\% across attacks. Clean accuracy of SAFT is lower than Full AT by 2.5\%, likely reflecting that the frozen parameters provide less flexibility for fitting clean examples.

Sparse update can outperform dense update in robustness. The observation that SAFT surpasses Full AT on ViT-B is particularly noteworthy. We attribute this to an implicit regularization effect. By restricting adversarial updates to a small sparse subspace, SAFT reduces the effective capacity of the optimizer and mitigates robust overfitting. This finding suggests that full parameter updates are not necessarily the upper bound for adversarial fine-tuning. SAFT also applies to CNNs. On ResNet-50 and CIFAR-10, it reaches 49.32\% PGD-20 accuracy compared to the accuracy of\ 48.95\% for Full AT (Appendix~\ref{app:resnet}).

\subsection{Main Results across Datasets}
\label{sec:main_dataset}

We further evaluate SAFT on CIFAR-100 and TinyImageNet to test whether 
the conclusions generalize to more complex datasets. Results are 
reported in Table~\ref{tab:datasets}, with CIFAR-10 in Table~\ref{tab:models}. SAFT's advantage over LoRA-based PEFT-AT grows with dataset complexity. On CIFAR-10, SAFT improves over LoRA by about 3\% on PGD-20 and AutoAttack. On CIFAR-100, the improvement is around 3\%. On TinyImageNet, the gap becomes dramatic. LoRA and LNLoRA collapse to low robust accuracy, while SAFT maintains 18.01\% on PGD-20 and 14.34\% on AutoAttack, an improvement of more than 11\% and 9\% respectively. The collapse of LoRA on TinyImageNet suggests that as the downstream task becomes more distant from the pretraining distribution, the low-rank optimization subspace becomes increasingly inadequate for capturing the directions needed for robustness. Sparse update, by contrast, draws from the full parameter space and remains effective even under significant distribution shift.

\begin{table}[h]
\centering
\caption{Comparison across datasets on ViT-B (CIFAR-10 results are in Table~\ref{tab:models}). SAFT consistently outperforms LoRA-based PEFT-AT in robust accuracy across CIFAR-10, CIFAR-100, and TinyImageNet. The robustness gap of LoRA-based methods grows with 
dataset complexity, while SAFT remains close to full-parameter AT.}
\label{tab:datasets}
\setlength{\tabcolsep}{6pt}
\renewcommand{\arraystretch}{1.0}
\begin{tabular}{ccccccc}
\toprule
Dataset & Method & Clean & PGD-20 & CW-20 & PGD-100 & AutoAttack \\
\midrule
\ifshowrev
\fi
\multirow{5}{*}{CIFAR-100} 
 & Pretrained      & 86.10 & 0.00  & 0.00  & 0.00  & 0.00  \\
 & Full Parameter  & 64.14 & 24.42 & 25.14 & 24.09 & 22.58 \\
 \cmidrule(l){2-7}
 & LoRA            & 60.78 & 22.25 & 22.21 & 21.81 & 19.76 \\
   & LNLoRA          & 61.05 & 21.96 & 22.01 & 21.52 & 19.80 \\
 & \textbf{SAFT}   & \textbf{62.60} & \textbf{24.84} & \textbf{24.80} & \textbf{24.45} & \textbf{21.80} \\
\midrule
\multirow{5}{*}{TinyImageNet} 
 & Pretrained      & 80.94 & 0.00  & 0.00  & 0.00  & 0.00  \\
 & Full Parameter  & 58.82 & 18.64 & 18.36 & 18.20 & 14.92 \\
 \cmidrule(l){2-7}
 & LoRA            & 55.53 & 6.71  & 7.27  & 5.97  & 4.89  \\
   & LNLoRA          & 56.03 & 7.92  & 8.41  & 7.27  & 5.73  \\
 & \textbf{SAFT}   & 54.40 & \textbf{18.01} & \textbf{16.86} & \textbf{17.79} & \textbf{14.34} \\
\bottomrule
\end{tabular}
\end{table}

SAFT remains close to full-parameter AT across all three datasets. On CIFAR-100, SAFT slightly outperforms Full AT on PGD-20 and PGD-100, with a small gap on AutoAttack. On TinyImageNet, SAFT closely matches Full AT in robust accuracy, with a remaining gap of 0.4--1.5\%. Clean accuracy of SAFT is somewhat lower than Full AT and, on TinyImageNet, also lower than LoRA-based methods, reflecting the constraint imposed by freezing over 95\% of the parameters under significant distribution shift. Even so, SAFT preserves most of Full AT's robustness across datasets of varying complexity while training only a small fraction of parameters.

\subsection{Robustness Critical Parameter under Different Attacks}
\label{sec:overlap}

A natural concern is whether the parameters identified by SAFT depend on the specific attack used to compute adversarial gradients. We test this by identifying RCPs separately under four 
different attacks (PGD-20, PGD-100, CW-20, and AutoAttack) on ViT-B / CIFAR-10, and compute the pairwise overlap between the resulting RCPs sets.

Results are reported in Table~\ref{tab:overlap}. The overlap is consistently high across all attack pairs, ranging from 82.82\% to 92.42\%. The two PGD variants (PGD-20 and PGD-100) yield the highest overlap of 92.42\%, indicating that the number of attack iterations has only a small effect on which parameters are identified as critical. Across structurally different attacks (e.g., CW-20 vs AutoAttack), the overlap remains above 82\%, far higher than the 5\% level expected if RCPs were arbitrary 5\% subsets.

Hence, adversarial gradient saliency does not overfit to a specific perturbation method, and the identified RCPs reflect a structural property of the model rather than an artifact of the attack used during selection. Section~\ref{sec:specificity} tests whether they are also robustness-specific.

\begin{table}[h]
\centering
\begin{minipage}[t]{0.5\textwidth}
\centering
\caption{Overlap ratio (\%) of RCPs under different attacks on ViT-B with CIFAR-10.}
\label{tab:overlap}
\small
\setlength{\tabcolsep}{4pt}
\renewcommand{\arraystretch}{1.1}
\begin{tabular}{lcccc}
\toprule
            & PGD-20 & PGD-100 & CW-20 & AA \\
\midrule
PGD-20      & --     & 92.42  & 85.87 & 83.88 \\
PGD-100     & 92.42  & --     & 86.90 & 87.96 \\
CW-20       & 85.87  & 86.90  & --    & 82.82 \\
AA          & 83.88  & 87.96  & 82.82 & --    \\
\bottomrule
\end{tabular}
\end{minipage}\hfill
\begin{minipage}[t]{0.46\textwidth}
\centering
\caption{Training with 5\% of parameters with different criteria on ViT-B with CIFAR-10.}
\label{tab:criteria}
\small
\setlength{\tabcolsep}{4pt}
\renewcommand{\arraystretch}{1.1}
\begin{tabular}{lcc}
\toprule
Selection criterion & Clean & PGD-20 \\
\midrule
Random                   & 52.32 & 11.92 \\
Weight magnitude         & 38.32 & 22.18 \\
Clean-data saliency      & 54.33 & 25.36 \\
Adv.\ saliency (SAFT)    & 82.55 & 44.41 \\
\bottomrule
\end{tabular}
\end{minipage}
\end{table}

\subsection{Specificity of Robustness Critical Parameters}
\label{sec:specificity}
RCPs might simply be generally important parameters. We test this on ViT-B and CIFAR-10 with three controls. \textbf{Set identity.} The top 5\% of parameters by pretrained weight magnitude share only 5.27\% of their entries with the RCPs, and the top 5\% by clean-data gradient saliency share 30.94\%. \textbf{Training under each criterion.} We train separate models that update only the 5\% selected by each criterion, with identical budget, schedule and seed (Table~\ref{tab:criteria}). Adversarial saliency beats clean-data saliency by 28.2 points in clean accuracy and 19.1 points on PGD-20 although the two selections overlap by a third, so the gain comes from the adversarial-specific part of the selection. \textbf{Zeroing.} Zeroing the RCPs of the trained SAFT model reduces PGD-20 accuracy by 86.0\% in relative terms, compared with 57.8\% for the weight-magnitude subset and 10.2\% for a random subset (Appendix~\ref{app:zeroing}). Together with the attack invariance in Table~\ref{tab:overlap}, these controls indicate that RCPs are specific to adversarial adaptation.

\begin{table}[!t]
\centering
\caption{SAFT under different AT objectives on ViT-B and CIFAR-10. SAFT consistently outperforms LoRA-based PEFT-AT under all three AT objectives.}
\label{tab:at_methods}
\setlength{\tabcolsep}{6pt}
\renewcommand{\arraystretch}{1.0}
\begin{tabular}{ccccccc}
\toprule
Method & AT Objective & Clean & PGD-20 & CW-20 & PGD-100 & AutoAttack \\
\midrule
\ifshowrev
\fi
Full Parameter      & TRADES  & 86.24 & 42.88 & 43.61 & 42.55 & 41.38 \\
LoRA                & TRADES  & 86.26 & 40.22 & 39.96 & 39.79 & 38.00 \\
LNLoRA              & TRADES  & 86.04 & 39.47 & 39.16 & 39.02 & 37.38 \\
\textbf{SAFT}       & TRADES  & \textbf{87.00} & \textbf{43.07} & \textbf{43.06} & \textbf{42.74} & \textbf{41.04} \\
\midrule
Full Parameter      & MART    & 75.91 & 48.73 & 46.12 & 48.57 & 44.73 \\
LoRA                & MART    & 66.66 & 42.31 & 39.37 & 42.30 & 38.00 \\
LNLoRA              & MART    & 66.47 & 42.17 & 38.67 & 42.04 & 37.31 \\
\textbf{SAFT}       & MART    & \textbf{70.68} & \textbf{45.12} & \textbf{41.39} & \textbf{45.07} & \textbf{40.50} \\
\bottomrule
\end{tabular}
\end{table}

\subsection{Generalization Across AT Objectives}
\label{sec:at_methods}

To test whether SAFT generalizes beyond the standard PGD-AT objective, we apply it to two additional AT objectives, TRADES~\cite{zhang2019theoretically} and MART~\cite{wang2019improving}, both designed to address the clean-robust trade-off through loss formulations. Results on ViT-B and CIFAR-10 are reported in Table~\ref{tab:at_methods}. SAFT outperforms LoRA-based methods under three AT objectives. The improvement under MART is large, where SAFT improves over LoRA by 4.02\% in clean accuracy and 2.81\% on PGD-20.

SAFT can surpass full-parameter AT under standard objectives. Under TRADES, SAFT also surpasses full-parameter AT on PGD-20 and PGD-100, and improves clean accuracy from 86.24\% to 87.00\%. Under MART, full-parameter AT achieves the highest robust accuracy overall, with SAFT trailing by 3.5--4.7\% across attacks. We attribute this to MART's strong reliance on per-example reweighting, which benefits from updating the full parameter set. Even so, SAFT remains the strongest PEFT-AT method under MART, improving over LoRA by 2.81\% on PGD-20 and 2.50\% on AutoAttack. 

\begin{wraptable}{r}{0.40\textwidth}
\centering
\vspace{-13pt}
\caption{Comparison with LoRA variants on ViT-B and CIFAR-10.}
\label{tab:lora_variants}
\footnotesize
\setlength{\tabcolsep}{3pt}
\renewcommand{\arraystretch}{1.0}
\begin{tabular}{lcccc}
\toprule
Method & Trainable & Clean & PGD-20 & AA \\
\midrule
LoRA          & 4.7M & 79.51 & 41.58 & 38.63 \\
LNLoRA        & 4.8M & 79.59 & 41.79 & 39.07 \\
DoRA          & 4.8M & 80.16 & 43.18 & 40.29 \\
\textbf{SAFT} & \textbf{3.1M} & \textbf{82.55} & \textbf{44.41} & \textbf{41.61} \\
\bottomrule
\end{tabular}
\vspace{-12pt}
\end{wraptable}

\textbf{Comparison with LoRA variants.} Table~\ref{tab:lora_variants} further compares SAFT with DoRA~\cite{liu2024dora}, a stronger LoRA variant, on ViT-B and CIFAR-10 under PGD-AT. DoRA improves over LoRA by 1.60\% on PGD-20 and 1.66\% on AutoAttack but still trails SAFT despite a larger trainable budget. Since its directional update remains rank-$r$ constrained, this indicates that the limitation of LoRA-based PEFT-AT lies in the low-rank constraint itself.

\subsection{Effect of Robustness Critical Parameter Percentage}
\label{sec:percentage}

\begin{wraptable}{r}{0.5\textwidth}
\centering
\vspace{-18pt}
\caption{Effect of the RCPs percentage on ViT-B and CIFAR-10. Even with only
1\% of parameters selected as RCPs, SAFT recovers most of the robustness of full-parameter AT.}
\label{tab:percentage}
\footnotesize
\setlength{\tabcolsep}{2.6pt}
\renewcommand{\arraystretch}{1.0}
\begin{tabular}{lccccc}
\toprule
RCPs \% & Clean & PGD-20 & CW-20 & PGD-100 & AA \\
\midrule
\ifshowrev
\fi
1\%               & 81.51 & 43.71 & 43.58 & 43.35 & 40.82 \\
3\%               & 81.56 & 43.88 & 44.01 & 43.60 & 41.20 \\
5\%               & 82.55 & 44.41 & 44.87 & 44.04 & 41.61 \\
7\%               & 82.66 & \textbf{44.71} & \textbf{44.87} & \textbf{44.29} & \textbf{41.68} \\
\midrule
100\% (Full AT)   & \textbf{85.04} & 43.60 & 44.57 & 42.96 & 41.16 \\
\bottomrule
\end{tabular}
\vspace{-12pt}
\end{wraptable}

To understand how robustness depends on the size of the selected RCPs subset, we vary the percentage from 1\% to 7\% on ViT-B / CIFAR-10 and compare with full-parameter AT (100\%). Results are shown in Table~\ref{tab:percentage}. With only 1\% of parameters selected, SAFT achieves 43.71\% on PGD-20 and 40.82\% on AutoAttack, already close to full-parameter AT. It indicates that the robustness-critical subspace is highly concentrated, and that even an extremely sparse update is sufficient to reach the robustness ceiling of full AT.

Increasing the budget from 1\% to 5\% yields modest additional gains (PGD-20 from 43.71\% to 44.41\%, AutoAttack from 40.82\% to 41.61\%), and clean accuracy improves by about 1\%. Beyond 5\%, robustness gains become marginal (PGD-20 changes from 44.41\% to 44.71\% between 5\% and 7\%, with CW-20 unchanged at 44.87\%). This fast saturation reinforces the sparsity finding in Fig.~\ref{fig:abs_gradient}.

\subsection{Training Cost and Further Analysis}
\label{sec:cost}\label{sec:further}

\begin{table}[b]
\centering
\begin{minipage}[t]{0.47\textwidth}
\centering
\caption{Measured training cost on ViT-B and CIFAR-10 on a single RTX 4090.}
\label{tab:cost}
\footnotesize
\setlength{\tabcolsep}{3pt}
\renewcommand{\arraystretch}{1.0}
\begin{tabular}{lccc}
\toprule
Method & Peak (GiB) & Step (s) & 40 ep.\ (h) \\
\midrule
Full AT          & 6.551 & 1.118 & 4.856 \\
LoRA ($r{=}32$)  & 6.129 & 1.629 & 7.078 \\
SAFT             & 5.582 & 1.114 & 4.840 \\
\bottomrule
\end{tabular}
\end{minipage}\hfill
\begin{minipage}[t]{0.47\textwidth}
\centering
{
\caption{Overlap (\%) between the RCPs recomputed at a given epoch and the calibration mask (ViT-B and CIFAR-10).}
\label{tab:drift}
\footnotesize
\setlength{\tabcolsep}{4pt}
\renewcommand{\arraystretch}{1.0}
\begin{tabular}{lccccc}
\toprule
Epoch & 1 & 10 & 20 & 30 & 40 \\
\midrule
Full AT & 95.65 & 50.97 & 51.45 & 50.08 & 50.03 \\
SAFT    & 95.65 & 55.73 & 56.42 & 56.75 & 57.58 \\
\bottomrule
\end{tabular}\par}
\end{minipage}
\end{table}

\textbf{Training cost.} Table~\ref{tab:cost} reports the training cost on ViT-B and CIFAR-10 under an identical configuration. SAFT lowers peak memory by 0.97\,GiB relative to Full AT and by 0.55\,GiB relative to LoRA. Its time per step matches that of Full AT, since all gradients are still computed, and is 32\% lower than that of LoRA. The 40-epoch times are extrapolated from the time per step. The parameter-side saving, at least $4kN(1-\rho)$ bytes for $N$ parameters and $k$ optimizer states per parameter, grows linearly with model size: under Adam it reaches 2.18 and 4.54\,GiB for ViT-L and ViT-H (Appendix~\ref{app:memory}). Activations dominate peak memory in our setting, so the share of the parameter side grows for larger backbones and smaller batches.

\textbf{Mask stability.} Recomputing adversarial saliency during training in Table~\ref{tab:drift} and Appendix~\ref{app:drift} shows that the RCP set drifts. Its overlap with the calibration mask falls from 95.65\% after epoch 1 to 55.73\% at epoch 10 and then stays stable, an order of magnitude above the roughly 5\% expected for random subsets. The static mask is thus a workable rather than an optimal choice.

\textbf{Distribution shift.} On CIFAR-10-C, CIFAR-100-C, Tiny-ImageNet-C~\cite{hendrycks2019benchmarking}, CIFAR-10.1~\cite{recht2019imagenet} and CINIC-10~\cite{darlow2018cinic}, SAFT achieves higher clean and PGD-20 accuracy than LoRA under every shift, attains the highest PGD-20 accuracy among all adapted models, and never suffers a larger relative accuracy drop than LoRA, as shown in Table~\ref{tab:shift_robust} and Appendix~\ref{app:shift}.

\begin{table}[h]
\centering
{
\caption{No-attack / PGD-20 accuracy (\%) on shifted data. Corruptions use severity 5. CINIC-10 IN uses a class-balanced 10,000-image subsample.}
\label{tab:shift_robust}
\setlength{\tabcolsep}{6pt}
\renewcommand{\arraystretch}{1.0}
\begin{tabular}{lcccc}
\toprule
Benchmark & Pretrained & Full AT & LoRA & SAFT \\
\midrule
CIFAR-10.1      & --           & 72.80 / 25.05 & 66.50 / 27.10 & 70.00 / 28.20 \\
CINIC-10 IN     & --           & 60.62 / 22.54 & 56.77 / 23.32 & 59.05 / 24.30 \\
CIFAR-10-C      & 77.01 / 0.00 & 69.58 / 25.21 & 65.50 / 26.79 & 68.29 / 27.86 \\
CIFAR-100-C     & 55.03 / 0.01 & 43.81 / 11.37 & 41.80 / 11.43 & 43.77 / 12.49 \\
Tiny-ImageNet-C & 35.20 / 0.00 & 19.50 / 2.57  & 16.99 / 0.88  & 17.73 / 2.79  \\
\bottomrule
\end{tabular}\par}
\end{table}

\section{Conclusion}
\label{Conclusion}
In this work, we revisit adversarial fine-tuning from a parameter-space perspective. Through gradient analysis, we find that adversarial optimization is concentrated in a sparse subset of parameters, which we call RCPs. A masking analysis further shows that this same subset is causally critical for the robust behavior of the trained model. Building on this finding, we propose SAFT, which identifies RCPs via adversarial gradient saliency and performs adversarial fine-tuning by updating only these parameters. Across 
three architectures, three datasets, and three AT objectives, SAFT consistently outperforms LoRA-based PEFT-AT and approaches or surpasses full-parameter AT in robustness while updating fewer than 5\% of parameters. It reduces optimizer-state, persistent-gradient and peak training memory. These results suggest that the optimization subspace relevant to adversarial robustness is sparse rather than low-rank, offering a parameter-space perspective complementary to existing loss-level approaches to AT.

\textbf{Limitations.} 
We acknowledge three main limitations. First, despite updating fewer than 5\% of parameters, SAFT does not bring wall-clock speedup or FLOP reduction over full-parameter AT. The current implementation uses unstructured sparsity, which reduces optimizer state and persistent gradient memory but does not accelerate forward or backward passes on standard hardware. Structured sparse selection at the head or row level is a promising direction to obtain real training speedup. Second, while SAFT improves the trade-off between trainable parameters and robustness, the clean-robust trade-off itself is not eliminated. Clean accuracy of SAFT remains below full-parameter AT in most settings. Third, the RCP mask is static, although about half of it drifts during training (Section~\ref{sec:further}); we have not evaluated periodic re-selection.

\section*{Acknowledgements}
This work was supported in part by the National Natural Science Foundation of China (No.~62376202). We declare no competing interests.

\bibliographystyle{abbrv}
\bibliography{refs}

%%%%%%%%%%%%%%%%%%%%%%%%%%%%%%%%%%%%%%%%%%%%%%%%%%%%%%%%%%%%

\appendix
\input{appendix_revision.tex}

%%%%%%%%%%%%%%%%%%%%%%%%%%%%%%%%%%%%%%%%%%%%%%%%%%%%%%%%%%%%

%\newpage
%\input{checklist.tex}

\end{document}

%% file: appendix_revision.tex
% ===================================================================
% Appendix added for the camera-ready version (all content is new and
% comes from the author rebuttal). Printed in red when \showrevtrue.
% ===================================================================
\begingroup

\newpage

\section{Memory Scaling of the Parameter Side}
\label{app:memory}
For $N$ parameters stored in fp32 and an optimizer with $k$ state tensors per parameter, full-parameter AT holds $4N$ bytes of weights, $4N$ of gradients and $4kN$ of optimizer state. SAFT holds the same weights and at most $4N$ of transient gradients, but optimizer state only for the trainable fraction $\rho$, i.e., at most $8N+4k\rho N$ bytes. The saving of at least $4kN(1-\rho)$ depends only on model size and optimizer order. Table~\ref{tab:memory_scaling} extrapolates it to larger backbones for Adam ($k=2$), using the trainable fraction of ViT-B ($\rho\approx3.6\%$). For SGD with momentum ($k=1$), which we use in our experiments, the saving is halved.

The parameter count grows as $d^2L$, whereas activation memory grows as $B\,T\,d\,L$ for batch size $B$, sequence length $T$, width $d$ and depth $L$. Our measurement in Table~\ref{tab:cost} (a large batch and only $T=65$ tokens at $32\times32$ resolution with patch size 4) is therefore the least favourable case for SAFT. For larger backbones, smaller batches and gradient checkpointing, the parameter side becomes the dominant term.

\begin{table}[h]
\centering
\caption{Analytic parameter-side memory (weights, gradients and Adam states, fp32) of Full AT and SAFT, extrapolated with the trainable fraction of ViT-B.}
\label{tab:memory_scaling}
\small
\setlength{\tabcolsep}{8pt}
\begin{tabular}{lcccc}
\toprule
Backbone & Params & Full AT (GiB) & SAFT (GiB) & Saved (GiB) \\
\midrule
ViT-B        & 85.2M & 1.27  & 0.66  & 0.61  \\
ViT-L        & 304M  & 4.53  & 2.35  & 2.18  \\
ViT-H        & 632M  & 9.42  & 4.88  & 4.54  \\
ViT-g/14     & 1.1B  & 16.39 & 8.49  & 7.90  \\
ViT-G/14     & 1.8B  & 26.82 & 13.90 & 12.92 \\
SwinV2-G     & 3.0B  & 44.70 & 23.16 & 21.54 \\
InternViT-6B & 5.9B  & 87.92 & 45.56 & 42.36 \\
\bottomrule
\end{tabular}
\end{table}

\section{Other Datasets and a Convolutional Backbone}
\label{app:resnet}
We repeat the gradient analysis of Section~\ref{sec:empirical_findings} on further datasets and on ResNet-50~\cite{he2016resnet}. Table~\ref{tab:concentration} reports the share of the cumulative adversarial saliency mass captured by the top 5\% of parameters. The concentration is consistent across datasets for ViT-B and weaker but still far above the uniform level of 5\% for the convolutional backbone. On ResNet-50 / CIFAR-10, SAFT reaches 49.32\% PGD-20 accuracy, compared with 48.95\% for Full AT. 

\begin{table}[h]
\centering
\caption{Share of cumulative adversarial gradient saliency captured by the top 5\% of parameters.}
\label{tab:concentration}
\small
\setlength{\tabcolsep}{8pt}
\begin{tabular}{llc}
\toprule
Backbone & Dataset & Top-5\% share (\%) \\
\midrule
ViT-B     & CIFAR-10     & 67.7  \\
ViT-B     & CIFAR-100    & 61.88 \\
ViT-B     & TinyImageNet & 63.59 \\
ResNet-50 & CIFAR-10     & 43.15 \\
\bottomrule
\end{tabular}
\end{table}

\section{Zeroing Interventions on the SAFT Model}
\label{app:zeroing}
Table~\ref{tab:mask} zeroes parameters of the Full AT model. Table~\ref{tab:zeroing} extends the intervention to the trained SAFT model on ViT-B and CIFAR-10 with two further criteria, each zeroing 5\% of the parameters. Zeroing the RCPs causes by far the largest damage, whereas the weight-magnitude subset causes smaller and differently structured damage, even though large weights are generally important.

\begin{table}[h]
\centering
\caption{Zeroing 5\% of the parameters of the trained SAFT model. Relative drops are computed with respect to the unmasked SAFT model (82.55\% clean, 44.41\% PGD-20).}
\label{tab:zeroing}
\small
\setlength{\tabcolsep}{8pt}
\begin{tabular}{lcccc}
\toprule
Zeroed subset (5\%) & Clean & PGD-20 & Rel.\ drop, clean & Rel.\ drop, PGD-20 \\
\midrule
Random           & 76.41 & 39.86 & 7.4\%  & 10.2\% \\
Weight magnitude & 62.69 & 18.75 & 24.1\% & 57.8\% \\
RCPs             & 30.60 & 6.23  & 62.9\% & 86.0\% \\
\bottomrule
\end{tabular}
\end{table}

\section{Drift of the Robustness-Critical Parameters during Training}
\label{app:drift}
We recompute adversarial gradient saliency at several epochs and compare each resulting top-$\rho$ set with the mask fixed at calibration, for the SAFT run and, as a reference, a Full AT run (Table~\ref{tab:drift}). The set reorganizes once as the model leaves its pretrained state and is stable thereafter. At the end of training it still overlaps the calibration mask by 57.58\%, an order of magnitude above the roughly 5\% expected for two random subsets of this size. Table~\ref{tab:percentage} further shows that reducing the budget from 5\% to 1\% costs only 0.70 points on PGD-20 and 0.79 on AutoAttack. This suggests that robustness is carried by a core substantially smaller than the mask, and that a static mask suffices as long as it contains this core. We did not evaluate dynamic re-selection.

\section{Distribution Shift}
\label{app:shift}
\textbf{Synthetic shifts.} We report the mean corruption accuracy (mCA) on CIFAR-10-C, CIFAR-100-C and Tiny-ImageNet-C~\cite{hendrycks2019benchmarking}, averaged over the 15 standard corruption types at all 5 severity levels, and the relative degradation $\Delta\% = (\text{Clean}-\text{mCA})/\text{Clean}\times100$, where Clean is the in-distribution accuracy of each model from Tables~\ref{tab:models} and~\ref{tab:datasets} (lower is better).
\textbf{Natural shifts.} We evaluate on CIFAR-10.1~\cite{recht2019imagenet}, an independently re-collected CSFATIFAR-10 test set, and on the ImageNet-sourced portion of the CINIC-10~\cite{darlow2018cinic} test split.
\textbf{Robustness under shift.} We additionally evaluate PGD-20 under both shift families. For corruptions we use severity 5 with a stratified sample of 1,000 images per corruption type. For CINIC-10 we use a class-balanced 10,000-image subsample of the same ImageNet-sourced pool, which is why its no-attack accuracies in Table~\ref{tab:shift_robust} differ slightly from Table~\ref{tab:shift_natural}.
SAFT has a smaller relative degradation than LoRA on every benchmark and than Full AT on three of the five, and the highest PGD-20 accuracy under every shift.

\begin{table}[h]
\centering
\caption{Mean corruption accuracy (mCA, \%) and relative degradation ($\Delta\%$, lower is better) on ViT-B.}
\label{tab:shift_synthetic}
\setlength{\tabcolsep}{6pt}
\begin{tabular}{lcccccc}
\toprule
 & \multicolumn{2}{c}{CIFAR-10-C} & \multicolumn{2}{c}{CIFAR-100-C} & \multicolumn{2}{c}{Tiny-ImageNet-C} \\
\cmidrule(lr){2-3}\cmidrule(lr){4-5}\cmidrule(lr){6-7}
Model & mCA & $\Delta\%$ & mCA & $\Delta\%$ & mCA & $\Delta\%$ \\
\midrule
Pretrained & 86.81 & 9.81  & 68.51 & 20.44 & 51.81 & 35.99 \\
Full AT    & 76.16 & 10.45 & 52.85 & 17.61 & 28.01 & 52.38 \\
LoRA       & 70.73 & 11.04 & 50.19 & 17.43 & 25.12 & 54.76 \\
SAFT       & 73.98 & 10.38 & 51.99 & 16.95 & 25.73 & 52.70 \\
\bottomrule
\end{tabular}
\end{table}

\begin{table}[h]
\centering
\caption{Accuracy (\%) and relative degradation ($\Delta\%$) under natural shift on ViT-B. CINIC-10 IN: ImageNet-sourced portion of the CINIC-10 test split.}
\label{tab:shift_natural}
\setlength{\tabcolsep}{6pt}
\begin{tabular}{lccccc}
\toprule
 & & \multicolumn{2}{c}{CIFAR-10.1} & \multicolumn{2}{c}{CINIC-10 IN} \\
\cmidrule(lr){3-4}\cmidrule(lr){5-6}
Model & Clean & Acc. & $\Delta\%$ & Acc. & $\Delta\%$ \\
\midrule
Full AT & 85.04 & 72.80 & 14.39 & 60.11 & 29.31 \\
LoRA    & 79.51 & 66.50 & 16.36 & 56.50 & 28.94 \\
SAFT    & 82.55 & 70.00 & 15.20 & 58.79 & 28.78 \\
\bottomrule
\end{tabular}
\end{table}

\section{Broader Impact}
\label{app:impact}
SAFT contributes to making adversarial training more accessible by reducing the optimizer-state and gradient memory cost of robust adaptation on large pretrained models, which lowers the barrier for researchers and practitioners with limited compute resources to deploy robust models in safety-sensitive applications. As a defense method, our work has no foreseeable negative societal impact beyond the general dual-use nature of adversarial robustness research.

\endgroup
\clearpage